\documentclass{article}

\usepackage{microtype}
\usepackage{graphicx}
\usepackage{subfigure}
\usepackage{booktabs} 

\usepackage{amssymb}

\usepackage{hyperref}

\usepackage[accepted]{icml2021}

\icmltitlerunning{Uncertainty Estimation and Out-of-Distribution Detection for Counterfactual Explanations: Pitfalls and Solutions}

\begin{document}

\twocolumn[
\icmltitle{Uncertainty Estimation and Out-of-Distribution Detection for Counterfactual Explanations: Pitfalls and Solutions}



\icmlsetsymbol{equal}{*}

\begin{icmlauthorlist}
\icmlauthor{Eoin Delaney}{ucd,insight,vm}
\icmlauthor{Derek Greene}{ucd,insight,vm}
\icmlauthor{Mark Keane}{ucd,insight,vm}
\end{icmlauthorlist}

\icmlaffiliation{ucd}{School of Computer Science, University College Dublin, Dublin, Ireland.}
\icmlaffiliation{insight}{Insight Centre for Data Analytics, Dublin, Ireland}
\icmlaffiliation{vm}{VistaMilk SFI Research Centre, Ireland}
\icmlcorrespondingauthor{Eoin Delaney}{eoin.delaney@insight-centre.org}

\icmlkeywords{XAI, Counterfactual Explanation, Counterfactual Evaluation}

\vskip 0.3in
]



\printAffiliationsAndNotice{}  

\begin{abstract}
Whilst an abundance of techniques have recently been proposed to generate counterfactual explanations for the predictions of opaque black-box systems, markedly less attention has been paid to exploring the uncertainty of these generated explanations. This becomes a critical issue in high-stakes scenarios, where uncertain and misleading explanations could have dire consequences (e.g., medical diagnosis and treatment planning). Moreover, it is often difficult to determine if the generated explanations are well grounded in the training data and sensitive to distributional shifts. This paper proposes several practical solutions that can be leveraged to solve these problems by establishing novel connections with other research works in explainability (e.g., trust scores) and uncertainty estimation (e.g., Monte Carlo Dropout). Two experiments demonstrate the utility of our proposed solutions.
\end{abstract}

\section{Introduction}
Predictions of opaque black-box systems are frequently deployed in high-stakes applications, such as finance, healthcare, and criminal justice \cite{adadi2018peeking}. Post-hoc counterfactual explanations can offer insights into the inner workings of black-box models and provide users with informative and actionable recourse \cite{Byrne2019, miller2019explanation}. Whilst an abundance of techniques have been proposed by the XAI community to generate counterfactual explanations \cite{keane2021if, karimi2021algorithmic}, significantly less attention has been paid to exploring the uncertainty of these explanations. The provision of uncertainty estimations on counterfactual explanations can avoid presenting users with overconfident and potentially harmful recourse and can improve decision-making, and build trust in intelligent systems \cite{bhatt2020uncertainty,jesson2020identifying}. Indeed, recent user studies \cite{mcgrath2020does} have demonstrated that people are more likely to agree with a model's prediction when given the corresponding predictive uncertainty, further motivating a need for solutions.  

Consider the example in a medical domain, where a system prescribes several prospective treatment plans to a patient suffering from respiratory problems in order to bolster their future lung capacity \cite{schulam2017reliable}. A black-box system considers the patient’s history and predicts the expected future lung capacity of the patient if they were prescribed a treatment plan. Formally, let $\mathbb{E}(Y_{A}| \mathcal{H})$ be the expected lung capacity of the patient under treatment $A$ given the history $\mathcal{H}$ of the patient. There may be a diverse set of prospective treatment plans available to the patient (recourse) and providing uncertainty estimates for the predictions can help medical practitioners to manage risk when prescribing a suitable treatment plan. Similar arguments for uncertainty estimations can also be made in other domains; for instance, exploring actions a store could take to boost future sales \cite{lucic2020does}. 

Motivated by such cases, it is clear that gaining insight into the uncertainty of suggested explanations is a key step in generating useful and trustworthy recourse, especially in high-stakes real-world prediction tasks \cite{molnar2020interpretable, upadhyay2021towards}. 

\subsection{Predictive Uncertainty and How it Occurs}
From a Bayesian perspective, total predictive uncertainty, ${\mathbb{V}(y|x)}$, can be decomposed into a sum of two components, namely epistemic (model) uncertainty and aleatoric (data) uncertainty \cite{kendall2017uncertainties, AWS_Uncertainty}. Let $x$ represent some input, $y$ the target variable and $\Theta$ the random parameters of the model, then  
\begin{equation}
    {\mathbb{V}(y|x)} = \underbrace{\mathbb{V}(\mathbb{E}(y|x,\Theta))}_{Epistemic} + \underbrace{{\mathbb{E}(\mathbb{V}(y|x,\Theta))}}_{Aleatoric}
\end{equation} 
\textit{Epistemic} uncertainty arises due to a lack of knowledge, typically stemming from a lack of training data \cite{gal2016uncertainty}. Explanations with low epistemic uncertainty are more likely under the data distribution. \textit{Aleatoric} uncertainty arises due to inherent noisiness, or stochasticity, in the data distribution. Instances close to the decision boundary can often be somewhat ambiguous, typically resulting in high aleatoric uncertainty \cite{schut2021generating}. Real-world scenarios are teeming with distributional shifts, cultivating epistemic uncertainty as the model faces new data for which it has less experience \cite{AWS_Uncertainty}. These shifts often compromise the validity of prescribed recourse, motivating the need for robust explanations \cite{rabanser2018failing,rawal2020can, upadhyay2021towards}. Indeed, the corresponding uncertainty estimates should also be sensitive and robust to these distributional shifts.

\section{Determining Counterfactual Uncertainty}
While a rich seam of research exists in uncertainty estimation, much of this work is relatively untapped, albeit extremely promising in the context of generating and evaluating explanations \cite{bhatt2020uncertainty}. Indeed quantifying uncertainty in counterfactual explanation is closely related to the task of ensuring that the generated explanations are plausible and not out-of-distribution (OoD). In this section we briefly survey related works, discussing some practical methods to quantify uncertainty in counterfactual explanation alongside the corresponding pitfalls. We suggest Trust Scores as a useful tool for this evaluation and test the utility of the discussed methods in subsequent experiments.

\textbf{Uncertainty as Prediction Probabilities}.
One baseline method of determining uncertainty in prediction (and explanation by analogy) is to consider the probability at the softmax or final layer of the black-box classifier as a proxy measure \cite{hendrycks2016baseline}. However, these probabilities are often poorly calibrated, resulting in a deterministically overconfident classification \cite{gal2016uncertainty, jiang2018trust, AWS_Uncertainty}. This problem represents a significant issue for algorithmic recourse for several reasons. Firstly, many popular counterfactual techniques explicitly incorporate these poorly calibrated softmax probabilities in optimizing to generate explanations \cite{wachter2017counterfactual, van2019interpretable}. Secondly, deterministic overconfidence, resulting from these poorly calibrated probabilities, can provide users with a false sense of trust in the system \cite{papenmeier2019model}. To tackle these difficulties, several calibration techniques have been developed to provide better estimates of predictive uncertainty than those provided by raw softmax probabilities, such as Temperature Scaling \cite{guo2017calibration}. Unfortunately, many of these approaches are not robust in capturing epistemic uncertainty that arises due to dataset drift \cite{ovadia2019can}.

\textbf{Leveraging Generative Models}. Many promote the use of variational auto-encoders (VAEs) and  generative adversarial networks (GANs) in counterfactual generation; the argument being that counterfactuals with low reconstruction errors should produce more realistic and less uncertain explanations \cite{mahajan2019preserving, kenny2020generating}. However, using GANs to detect out-of-distribution instances by measuring the likelihood under the data distribution can fail \cite{nalisnick2018deep}, while VAEs often generate ambiguous and blurry explanations. More recently, some researchers have argued that using auxiliary generative models in counterfactual generation incurs an engineering overhead and is not feasible for complex datasets, and generation through implicit minimization of epistemic and aleatoric uncertainties is more practical \cite{schut2021generating}. 

Monte Carlo dropout \cite{gal2016dropout} was developed as a Bayesian solution to uncertainty estimation in deep neural networks. To the best of our knowledge only one study has indicated the promise of this technique in evaluating uncertainty in counterfactual explanations, without the computationally expensive need to retrain a network or build an ensemble \cite{kenny2020generating}. Alternatively, deep ensembles \cite{lakshminarayanan_simple_2017} have enjoyed immense promise in estimating uncertainty in prediction and explanation \cite{schut2021generating}. However, one shortcoming of these methods is that they make assumptions about the to-be-explained model (e.g. MC-Dropout relies on the availability of dropout layers).

\textbf{Grounding Explanations in the Training Data.}
Despite a lack of clarity in how best to computationally evaluate counterfactual explanations \cite{keane2021if}, many agree that `good' counterfactual explanations should be grounded in the training data \cite{laugel2019dangers, keane2020good}. Case-Based Reasoning (CBR) methods have enjoyed notable success in generating counterfactual explanations. Leveraging the closest instance to the to-be-explained query that is in a different class (the so called \textit{nearest unlike neighbour} \cite{nugent2005case}) has demonstrated significant promise in generating counterfactual explanations for tabular \cite{keane2020good}, image \cite{goyal2019counterfactual}, and time series data \cite{delaney2020instance}. Other techniques have enjoyed success through harnessing instances in the training data that are maximally representative of a class (i.e., class prototypes) to guide counterfactual generation \cite{van2019interpretable}.  Indeed, monitoring proximity to the to-be-explained instance in terms of some $\ell{p}$ norm can be a simple but useful heuristic for determining feasibility \cite{karimi2020model}. However, solely minimizing the distance between an instance $x$ and a counterfactual $x'$ can ignore the data manifold and might prescribe recourse along an infeasible path \cite{poyiadzi2020face}. Building on the success of linear programming techniques in algorithmic recourse \cite{ustun2019actionable}, DACE \cite{kanamori2020dace} considers the empirical distribution when generating counterfactual explanations. In order to achieve this, Local Outlier Factor (LOF) scores \cite{breunig2000lof} can be incorporated into a cost function which is minimized using mixed-linear optimization to produce realistic counterfactual explanations. However, LOF is an unsupervised method and does not consider class labels, which are often readily available when evaluating post-hoc counterfactual explanations, motivating alternative solutions.  

\textbf{The Promise of Trust Scores}.
\textit{Trust scores} \cite{jiang2018trust} measure the ratio between (i) the distance from the testing sample to the nearest class different from the predicted class, and (ii) the distance to the predicted class. Thus, they capture the agreement between the classifier and a modified nearest-neighbor classifier on the testing example. A trust score of 1 means that the distance to the predicted class is the same as the distance to the nearest other class. High trust scores are preferable and theoretically correspond to a high probability of agreement with the Bayes optimal classifier \cite{jiang2018trust}.  We argue that trust scores should be useful in counterfactual evaluation for several reasons. Firstly, a key challenge here is providing explanations that are robust to distributional shifts \cite{upadhyay2021towards}. Trust scores were previously found to be useful in monitoring classifiers to detect distribution shifts, a key contributor to epistemic uncertainty \cite{jiang2018trust, de2021trust}. Secondly, trust scores are model agnostic and flexible across domains. Trust scores rely solely on the training data and make no assumptions about the model architecture, differentiability of an objective function, or the availability of an auto-encoder during training. They can also be extended to regression tasks \cite{de2021trust}. This readily facilitates the use of trust scores in comparative experiments, which has been a stumbling block for the XAI research community \cite{keane2021if}. Finally, well-maintained and well-documented open source software is available to compute trust scores \cite{klaise2020alibi}.

\section{Experiments}
In this section, we report two experiments that demonstrate the promise of quantifying uncertainty in counterfactual evaluation. In Experiment 1, we test trust scores in the context of detecting out-of-distribution instances and measuring epistemic uncertainty that arises from distributional shift. In Experiment 2, we explore the promise of uncertainty estimation when comparing counterfactual explanations that are produced by different techniques.

\textbf{Experiment 1: Testing Trust Scores.} In this experiment we explore the utility of trust scores in the context of detecting out-of-distribution (OoD) instances and quantifying predictive uncertainty. To emulate distributional shift we train a deep CNN on MNIST and then evaluate the predictions when the model is tested on data from; (i) MNIST, (ii) FashionMNIST (OoD). We would expect our evaluation metrics to deem images from the OoD FashionMNIST test set to be highly uncertain. Both datasets are formatted identically in terms of image size and grey-scale color. To quantify uncertainty, we monitor the softmax probability outputs of the network as a simple baseline \cite{hendrycks2016baseline}, and implement Monte Carlo dropout \cite{gal2016dropout} with 100 forward passes to compute the posterior mean (MC-Mean; higher is better) and posterior standard deviation (MC-Std; lower is better) of the predictions. Following \citet{kanamori2020dace}, we also implement 10-LOF \cite{breunig2000lof} to detect OoD instances.

\begin{figure}[!t]
\centering
\includegraphics[width=0.40\textwidth]{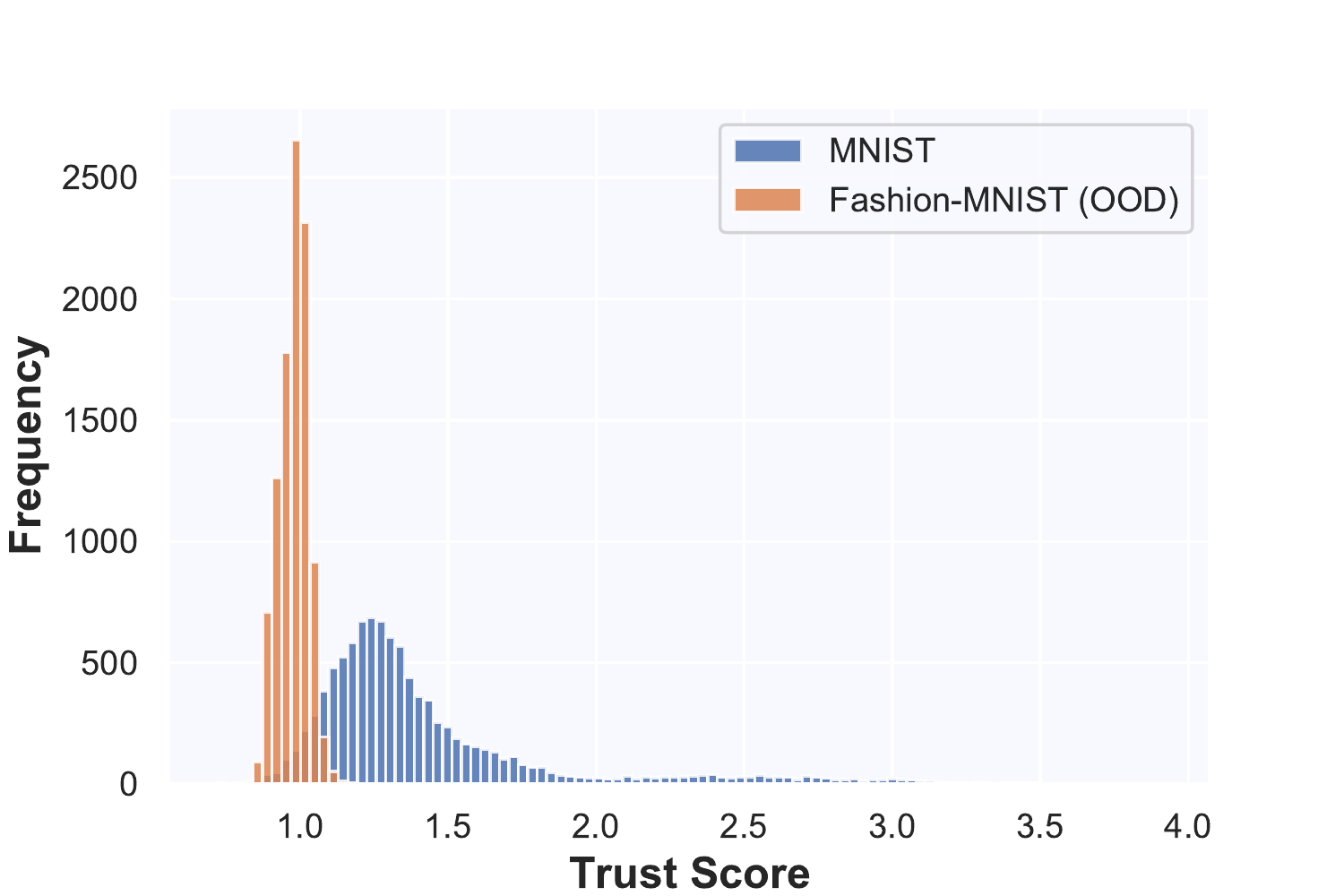}
 \caption{Comparing the distribution of Trust Scores for the test sets of (i) MNIST and (ii) FashionMNIST, CNN trained on MNIST.}
\end{figure}

\begin{figure*}[t!]
  \centering
  \subfigure[\textbf{\centering{Query}}]{\includegraphics[scale=3.3]{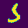}}
  \subfigure[\textbf{W-CF}]{\includegraphics[scale=3.3]{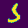}}
  \subfigure[\textbf{Proto-CF}]{\includegraphics[scale=3.3]{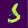}}
  \subfigure[\textbf{NUN-CF}]{\includegraphics[scale=3.3]{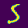}}
  \caption{Comparing Post-Hoc Counterfactual Explanations on MNIST for a Query Image: \textbf{True Label = 5} \& \textbf{Predicted Label = 3}. Three counterfactual explanations are of the form; ``\textit{If the image looked like this, the system would have correctly classified it as a 5}". The counterfactual explanation produced by Proto-CF is quite uncertain (MC-Mean = 0.56, MC-STD = 0.30, Trust Score = 1.00), relative to the explanation produced by NUN-CF (MC-Mean = 0.99, MC-STD = 0.03, Trust Score = 1.18). See supplement for additional examples.}
\end{figure*}

\textbf{Results.} Instances from the OoD FashionMNIST test set were found to have significantly lower trust scores compared to those from the original MNIST test set (Wilcoxon $p<0.01$), see Figure 1. Moreover, instances that were determined to be OoD by 10-LOF had significantly lower trust scores (mean = 0.971 $\pm$ 0.001) than those that were deemed to be within the data distribution (mean = 1.319 $\pm$ 0.004). These results indicate that trust scores can operate as a good proxy for OoD detection. Worryingly, many of the OoD instances were found to have high softmax probabilities, even if they had low-trust scores. For example, the CNN would often be $\approx 99\%$ sure an image of a shirt was actually an image of an eight, confirming that the softmax probabilities are a poor heuristic for detecting distributional shifts, whilst trust scores were more reliable. There is a strong monotonic relationship between MC-Mean and Trust Scores $r \approx 0.78$, indicating that trust scores are a useful proxy for uncertainty estimation in explanation. However, user studies will ultimately be needed to confirm these findings.

\textbf{Experiment 2: Comparing Counterfactuals.}
In our next experiment, we generate and comparatively evaluate counterfactual explanations for misclassifications of a CNN classifier on MNIST. We monitor uncertainty in the explanation using MC-Dropout and compute the Trust Scores of the generated explanations. Where possible, we also visualise the explanation. We evaluate the explanations produced by several popular counterfactual techniques that can readily generate explanations for three different data types (tabular, image, and time series):

\begin{itemize}
    \item \textbf{NUN-CF} \cite{nugent2005case}: This simple method retrieves the nearest neighbour to the query in a counterfactual class (nearest-unlike-neighbour). 
    \item \textbf{W-CF}: Inspired by \citet{wachter2017counterfactual}, this perturbation-based technique aims to generate sparse and proximal explanations by emulating the \textit{closest possible world}.  
    \item \textbf{Proto-CF}: Originally proposed by \citet{klaise2020alibi}, this state-of-the-art method facilitates the implementation of an auto-encoder to aid the generation of plausible counterfactuals by maximizing the likelihood of $x'$ under the training data distribution, following \citet{dhurandhar2018explanations}. Class prototypes are used to guide and speed up the counterfactual generation process.
\end{itemize}
\textbf{Results.} We observe that explanations produced by Proto-CF \cite{van2019interpretable} in the image domain are often quite ambiguous and blurry, confirming previous findings in other works \cite{schut2021generating, kenny2020generating}. The explanations are also quite uncertain according to MC-Dropout (MC-Mean $\approx$ $0.67$) and have relatively low trust scores (mean = 0.977 $\pm 0.008$), indicating that they poorly resemble instances in the training data of the counterfactual class. 

W-CF produces explanations with almost identical trust scores to the originally misclassified instances. This is somewhat unsurprising as W-CF is prone to generating adversarial examples \cite{wachter2017counterfactual}, frequently resulting in explanations that are extremely similar to the test image, which are often not perceptibly different due to small pixel-level changes \cite{kenny2020generating} (see Fig.2). 

On the other hand, explanations produced by NUN-CF are inherently well grounded in the training data (MC-Mean $\approx$ 0.93). Despite their simplicity, NUN-CF generates surprisingly good counterfactuals for MNIST misclassifications at the cost of sparsity and proximity. However, it is unclear how sparse or proximal good counterfactual explanations should be due to a lack of adequate user testing \cite{keane2021if}. Indeed, we note that explanations produced purely by NUN-CF are often not practical (e.g., when images are poorly aligned or less diverse training sets are available). However, such instances have been successfully used to guide counterfactual generation beyond MNIST in more complex color image datasets \cite{goyal2019counterfactual}, an inherently difficult problem for the XAI community.  

\section{Recommendations and Discussion}
Providing uncertainty estimations on counterfactual explanations is a relatively unexplored yet immensely promising problem that can greatly aid the provision of trustworthy recourse \cite{bhatt2020uncertainty, schut2021generating}. In light of this, we explore several practical solutions for quantifying uncertainty in counterfactual explanations. When explaining the predictions of neural networks we recommend using Monte Carlo dropout as a fast and efficient tool for determining uncertainty in explanation \cite{gal2016dropout, kenny2020generating}. We propose trust scores \cite{jiang2018trust} as a practical tool for evaluating how well explanations are grounded in the training data. Experiments demonstrate that trust scores provide a good proxy measure for uncertainty and for out-of-distribution detection. Moreover, unlike other popular techniques, trust scores make no assumptions about the model architecture or the availability of an auto-encoder and can also be readily extended to regression tasks \cite{de2021trust}. For a deeper exploration of the promise of uncertainty estimation in XAI, we recommend the survey compiled by \citet{bhatt2020uncertainty} and the excellent technical report by \citet{AWS_Uncertainty}.

Extending these experiments to more complex image and time series datasets and exploring the role of adversarial training in providing robust recourse and uncertainty estimates is an interesting avenue for future work \cite{upadhyay2021towards, schut2021generating}. Motivated by the lack of user studies in counterfactual evaluation \cite{keane2021if}, it will be important to conduct extensive user tests to explore what users deem to be out-of-distribution and what computational proxy might best capture such cases.



\section*{Acknowledgements}
This publication has emanated from research conducted with
the financial support of (i) Science Foundation Ireland (SFI)
to the Insight Centre for Data Analytics under Grant
Number 12/RC/2289\_P2 and (ii) SFI and the Department of
Agriculture, Food and Marine on behalf of the Government
of Ireland under Grant
Number 16/RC/3835 (VistaMilk).

\bibliography{example_paper}
\bibliographystyle{icml2021}

\clearpage
\appendix
\section{Supplementary Material}
\paragraph{Black-box model.}
For both experiments we train a convolutional neural network (CNN) as our black-box classifier following the architecture implemented by \citet{van2019interpretable}. Both MNIST \cite{lecun-mnisthandwrittendigit-2010} and FashionMNIST \cite{xiao2017fashion} images are scaled to [-0.5,0.5] and the default training and test sets are used. Dropout layers are implemented for regularization and conveniently facilitate uncertainty computations using MC-Dropout. We train with an Adam optimizer for 10 epochs using a batch size of 256. The classifier achieves an accuracy of 98.93\% on the MNIST test set leaving 107 to-be-explained images which are misclassified.  

For a black-box classifier $b$, let $x$ be some to-be-explained instance with predicted class $y$, and $x'$ be some candidate counterfactual explanation such that $b(x') = y'$.

\textbf{Softmax Probability.} \citet{hendrycks2016baseline} suggest that as a simple baseline for OoD detection and uncertainty estimation in deep neural networks is to monitor the softmax probability i.e monitor $p(y|x, D)$ where D is the training distribution. For counterfactual explanations this amounts to determining  $p(y'|x', D)$ \cite{schut2021generating}.

\textbf{Monte Carlo Dropout.} We provide a brief overview of how MC-Dropout \cite{gal2016dropout} works, closely following the description by \citet{AWS_Uncertainty}. Once a predictive distribution $p(y|x, D)$ is obtained, the corresponding uncertainty can be uncovered by exploring the variance. In order to learn this distribution we can learn the parametric posterior distribution p($\Theta|D)$ (i.e., the distribution over the model parameters).  

\citet{gal2016dropout} discovered that, by randomly switching off neurons in a neural network using different dropout configurations, one could approximate the parametric posterior distribution without the need to retrain the network. Each dropout configuration $\Theta_{t}$ corresponds to a sample from the approximate parametric posterior distribution $q(\Theta|D)$ s.t. $\Theta_{t} \sim q(\Theta|D)$. 

Sampling from the approximate posterior enables us to uncover the predictive distribution $p(y|x)$:
\begin{equation}
     p(y|x, D) \approx   \int_\Omega \underbrace{p(y|x, \Theta)}_{likelihood} \underbrace{q(\Theta|D}_{posterior}) \,d\Theta
\end{equation}
The above integral can be approximated through Monte Carlo methods, yielding; 
\begin{equation}
     p(y|x ,D) \underbrace{\approx}_{MC}   \frac{1}{T}\sum_{t=1}^{T} p(y|x, \Theta_{t})
\end{equation}
Indeed multiple forward passes with different dropout configurations allow us to uncover the predictive distribution. If we assume that likelihood is Gaussian distributed, the mean $f(x, \theta)$ and variance $s^{2}(x, \Theta)$ parameters of the Gaussian function are determined from the Monte Carlo simulations s.t $f(x, \theta), s^{2}(x, \Theta) \sim$ MC-Dropout(x), and can provide information about the predictive uncertainty through linking back with Equation 1. 
\begin{equation}
    {\mathbb{V}(y|x)} = \underbrace{\mathbb{V}(\overbrace{\mathbb{E}(y|x,\Theta)}^{f(x, \theta)})}_{Epistemic} + \underbrace{{\mathbb{E}(\overbrace{\mathbb{V}(y|x,\Theta)}^{s^{2}(x, \Theta)})}}_{Aleatoric}
\end{equation} 

\begin{figure}[!t]
\centering
\includegraphics[width=0.25\textwidth]{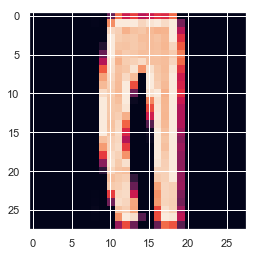}
 \caption{\textbf{Softmax Deterministic Overconfidence}; The black-box CNN classifier trained on MNIST images is 99.21 \% sure this image of `trouser' is an image of a `7'. Trust Score = 0.907.}
\end{figure}

\paragraph{Computing Trust Scores.}
We compute trust scores using the implementation provided by \cite{klaise2020alibi}. We set $k=10$, so that 10 - nearest neighbours are considered for distance calculations. We set the distance type to $point$ such that the distance to the $k$-th nearest neighbour is computed when determining the trust score. A Euclidean distance metric is used, we set the filter parameter $\alpha$ = 0 with a leaf size of 40 for kd-trees. 

While trust scores work best for low-mid dimensional datasets, the authors note that reducing dimensionality of complex datasets has negligible implications on the information conveyed by Trust Scores and can speed up computation \cite{jiang2018trust}. We compute the trust scores for the images without any dimensionality reduction so that they can fairly be considered with MC-Dropout. 

\textbf{Interpreting Trust Scores.}
A trust score of 1 means that the distance to the predicted class is the same as the distance to the nearest other class. One way of interpreting why the OoD FashionMNIST images have trust scores close to one (mean = 0.971 $\pm$ 0.001) is because they equally do not particularly resemble any of the classes in the MNIST training set. Unsurprisingly, the trust scores of the MNIST class prototypes are much higher (mean = 1.56 $\pm$ 0.103).

\textbf{Local Outlier Factor Method.}
The Local Outlier Factor Method \cite{breunig2000lof} was implemented as a novelty detector setting $k=10$ and using a Euclidean distance metric.  

\paragraph{IM1 Omission.}
IM1 \cite{van2019interpretable} is a popular metric for evaluating the realism of counterfactual explanations, based on the reconstruction losses of auto-encoders. Realism can be linked to epistimitic uncertainty \cite{schut2021generating}. We omit this metric from our evaluation for several reasons. Firstly, the metric requires separate auto-encoders to be trained on the to-be-explained and counterfactual class. For MNIST this would require training 10 separate auto-encoders that can successfully reconstruct the class (a non-trivial task). This is even more intractable for more complex datasets with more classes requiring even more autoencoders to be trained, where reconstruction of complex color images is much more difficult. 

\subsection{Counterfactual Methods}
In this section we provide more information about the counterfactual methods used in Experiment 2.

\textbf{Proto-CF.} Originally proposed by \citet{van2019interpretable}, this method aims to generate a counterfactual by minimizing a multi-objective loss function;
\begin{equation}
    Loss = c L_{pred} + \beta L_{1} + L_{2} + L_{AE} + L_{proto}
\end{equation}
The first term in the loss function encourages the perturbed instance to belong to the counterfactual class. The elastic net regularizer $\beta L_{1} + L_{2}$ aims to ensure sparsity and proximity in the generated instance. $L_{AE}$ is the reconstruction error of the candidate counterfactual instance which is minimized to encourage the counterfactual to belong to the training data distribution. However, to specifically guide the instance towards the distribution of the perturbed class the $L_{2}$ distance between the instance and the counterfactual class prototype is minimized in the $L_{proto}$ term.     

Following the recommendations of the authors we set hyperparameters; $\gamma$ = 100, $\Theta$ = 100, $c_{init}$ = 1, $c_{steps}$=2, and max iterations = 1000. We train a convolutional autoencoder following the architecture of \citet{van2019interpretable} to facilitate the use of the $L_{AE}$ term.

\textbf{W-CF.}
Inspired by \citet{wachter2017counterfactual}, this technique aims to generate a counterfactual by emulating the closest possible world and is implemented in \cite{klaise2020alibi}. The counterfactual instance $x'$ is generated through minimizing a simple loss function: 
\begin{equation}
    Loss = L_{pred} + \lambda L_{dist}
\end{equation}
We set hyperparameters; target class = $other$, target proba = 0.5 to emulate an instance close to the decision boundary or \textit{possible world}, tol = 0.01, $lam_{init}$ = 0.1, $lam_{steps}$=10, and max iterations = 1000. The Manhattan distance between the query and counterfactual is minimized in the loss function to generate sparse and proximal explanations.

\textbf{NUN-CF.} Originally proposed by \citet{nugent2005case}, this technique retrieves an explanation by locating the nearest neighbour in a counterfactual class. Alternatively, the target counterfactual class can be specified, which speeds up counterfactual retrieval as we consider a smaller pool of training images. Following \citet{kenny2020generating}, we use a 1-nearest neighbour search across the pixel space using $L_{2}$ distance to retrieve neighbours.

\textbf{Tabulated Results from Experiment 2.}
\begin{table}[h!]
\caption{The average performance of counterfactual explanations generated by different techniques for MNIST misclassifications (107 in total) according to MC-Mean (higher is better), MC-STD (lower is better) and Trust Scores.}
\label{sample-table}
\vskip 0.15in
\begin{center}
\begin{small}
\begin{sc}
\begin{tabular}{lcccr}
\toprule
Method & MC-Mean & MC-STD & Trust Score \\
\midrule
Proto-CF    & 0.667 & 0.242 &0.977 \\
W-CF & 0.761 & 0.294& 1.017\\
NUN-CF    & 0.931 & 0.115& 1.180 \\

\bottomrule
\end{tabular}
\end{sc}
\end{small}
\end{center}
\vskip -0.1in
\end{table}

\clearpage
\begin{figure*}[t!]
  \centering
  \subfigure[\textbf{\centering{Query}}]{\includegraphics[scale=5]{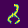}}
  \subfigure[\textbf{Proto-CF}]{\includegraphics[scale=5]{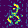}}
  \subfigure[\textbf{NUN-CF}]{\includegraphics[scale=5]{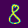}}

  \centering
  \subfigure{\includegraphics[scale=5]{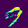}}
  \subfigure{\includegraphics[scale=5]{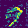}}
  \subfigure{\includegraphics[scale=5]{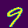}}
  
  \centering
  \subfigure{\includegraphics[scale=5]{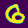}}
  \subfigure{\includegraphics[scale=5]{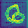}}
  \subfigure{\includegraphics[scale=5]{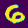}}
  
  \centering
  \subfigure{\includegraphics[scale=5]{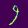}}
  \subfigure{\includegraphics[scale=5]{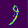}}
  \subfigure{\includegraphics[scale=5]{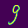}}
  \caption{Comparing counterfactual explanations generated by different methods for MNIST misclassifications.}
\end{figure*}


\end{document}